\documentclass[letterpaper, 10 pt, conference]{ieeeconf}  % Comment this line out if you need a4paper

\IEEEoverridecommandlockouts                              % This command is only needed if 
\usepackage{cite}
\usepackage{graphics} % for pdf, bitmapped graphics files
\usepackage{amsmath} % assumes amsmath package installed
\usepackage{amssymb}  % assumes amsmath package installed
\usepackage[T1]{fontenc}
\usepackage{graphicx}
\usepackage{xcolor}
\usepackage{bbm}
\usepackage{balance}
\usepackage{booktabs}
\usepackage{multirow}
\usepackage{pifont}
\usepackage{url}
\usepackage{eso-pic}
 
\title{\LARGE \bf
DPNeXt: A Lightweight Multi-Scale Feature Fusion Framework for Efficient ViT-Based Multi-Task Dense Prediction
}
\author{Jehun Kang$^{1}$, Jungha Wang$^{1}$, Youngjun Hwang$^{1}$, and David Hyunchul Shim$^{1\dagger}$%
\thanks{$^{\dagger}$Corresponding author.}%
\thanks{This work was supported by the Technology Innovation Program (RS-2025-25453109, Development of advanced synthetic data collection and simulation technology for autonomous driving) funded by the Ministry of Trade, Industry and Energy (MOTIE, Korea).}%
\thanks{$^{1}$The authors are with the School of Electrical Engineering, Korea Advanced Institute of Science and Technology (KAIST), Daejeon, South Korea.
        {\tt\small \{wpgnssla34, marco8522, yj.hwang, dhcshim\}@kaist.ac.kr}}%
}

\begin{document}

\AddToShipoutPicture*{%
  \AtPageLowerLeft{%
    \raisebox{0.32in}{%
      \hspace{0.55in}%
      \parbox{7.4in}{\centering\scriptsize
      \textcopyright~2026 IEEE. Personal use of this material is permitted.
      Permission from IEEE must be obtained for all other uses, including
      reprinting/republishing this material for advertising or promotional
      purposes, creating new collective works, for resale or redistribution
      to servers or lists, or reuse of any copyrighted component of this
      work in other works.}%
    }%
  }%
}

\maketitle
\thispagestyle{empty}
\pagestyle{empty}

\begin{abstract}
Multi-Task Learning (MTL) in robotics perception systems supports comprehensive 3D spatial scene understanding by integrating semantic segmentation and depth estimation. While Vision Foundation Models (VFMs) are increasingly adopted as robust feature encoders, existing decoding strategies present a critical bottleneck. To address this, we propose DPNeXt, a streamlined multi-scale feature fusion decoder and efficient alternative to the standard Dense Prediction Transformer (DPT). DPNeXt uses dual depthwise separable inverted bottlenecks to improve frozen VFM utilization through fusion-centric decoding and independent task modularization. To further mitigate negative inductive transfer between tasks, we introduce the Multi-Task Boundary Guidance (MTBG) strategy. Unlike prior boundary-aware methods that add fusion modules or gating, MTBG applies symmetric boundary-focused supervision to encourage geometric consistency without extra annotation or inference cost. Experiments on Cityscapes show that DPNeXt-S outperforms prior state-of-the-art (SOTA) MTL models, while DPNeXt-B further improves the overall performance and achieves the best results among the compared methods. On NYUv2, DPNeXt-B also achieves the best semantic segmentation and depth estimation results among the compared methods while requiring substantially fewer trainable parameters than prior large-scale MTL models. Compared with the standard DPT, DPNeXt-S reduces trainable parameters by 78.6\% and achieves the fastest inference speed among the compared models on resource-constrained laptop hardware. The source code, model checkpoints, and a demo video will be made available at \url{https://github.com/kangjehun/DPNeXt}.
\end{abstract}

%%%%%%%%%%%%%%%%%%%%%%%%%%%%%%%%%%%%%%%%%%%%%%%%%%%%%%%%%%%%%%%%%
\section{INTRODUCTION}

Holistic 3D scene understanding is crucial for downstream robotic tasks. While contemporary Bird's-Eye-View (BEV) \cite{li2024bevformer, chen2024vadv2} and Occupancy Grid \cite{tong2023scene} approaches effectively extract spatial information, their deployment is often constrained by expensive hardware such as 3D LiDAR and high-maintenance HD-Maps for training-label acquisition. In contrast, dense prediction offers a practical alternative by estimating pixel-level semantics and depth. Recently, Vision Foundation Models (VFMs) \cite{ravi2025sam, yang2024depthv2} have enabled the extraction of high-quality pseudo-labels from monocular cameras, making it highly scalable.

These dense representations can be used explicitly for 3D reconstruction or BEV cost maps. Alternatively, dense prediction can serve as an auxiliary training objective, where prediction heads are discarded or retained solely for debugging after training. The resulting fused latent features provide compact semantic and geometric cues that can support downstream perception modules.

\begin{figure}[t]
    \centering
    \includegraphics[width=\columnwidth]{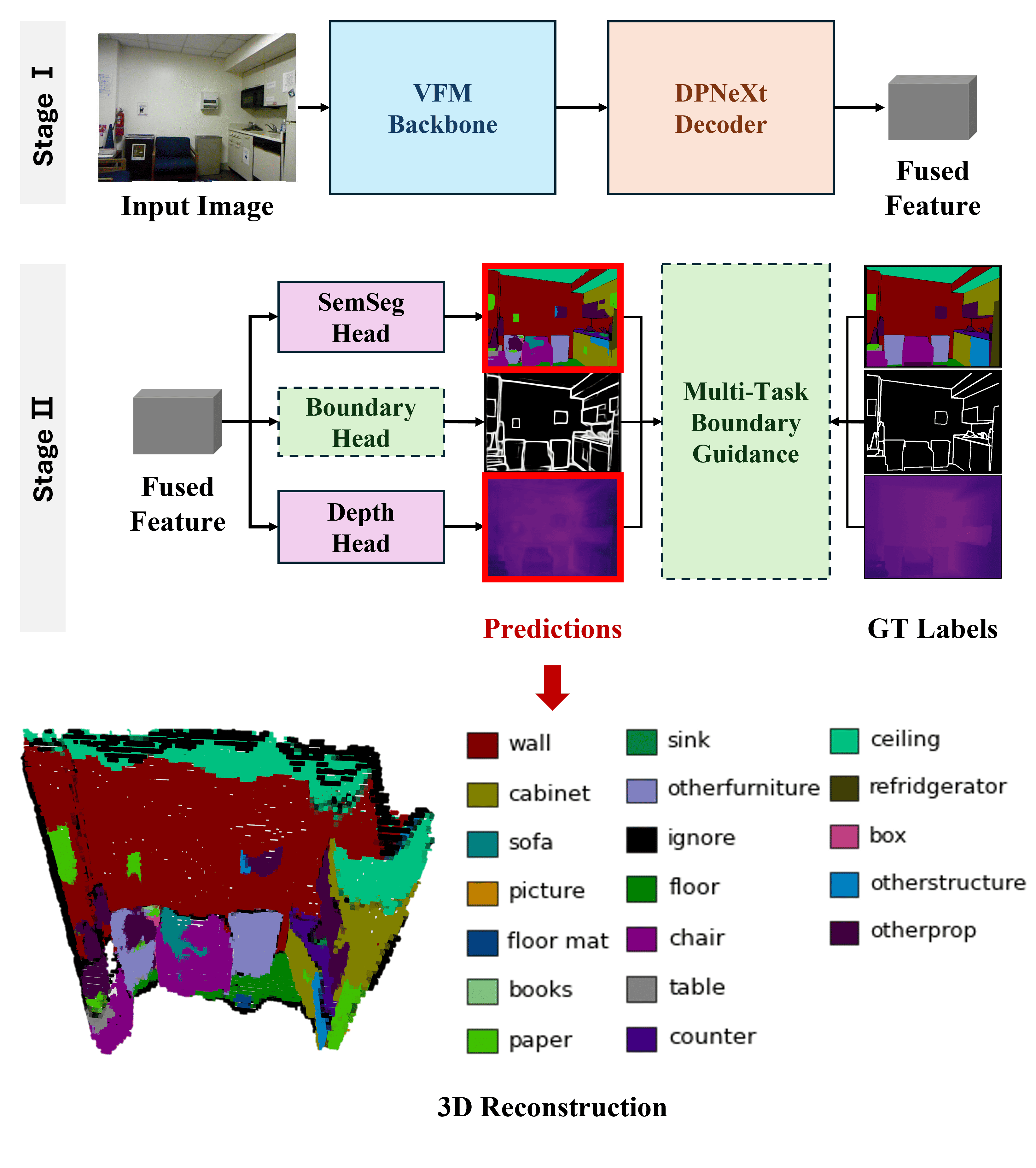}
    \caption[Proposed MTL framework.]{\textbf{Overview of the proposed MTL framework.} The architecture operates in two main stages. In Stage I, a frozen VFM backbone (\textcolor{blue}{Blue}) extracts rich representations, which the DPNeXt decoder (\textcolor{orange}{Orange}) aggregates into a shared fused feature (\textcolor{gray}{Gray}) through multi-scale fusion. This feature aggregates complementary multi-scale cues and can serve as a compact latent representation for downstream perception modules. In Stage II, task-specific heads process this shared feature, and are optimized during training by the Multi-Task Boundary Guidance strategy (\textcolor{green}{Green}) to encourage structural consistency and mitigate negative transfer. The final dense predictions (\textcolor{red}{Red}) can be explicitly utilized, as demonstrated by the 3D reconstruction example.}
    \label{fig:framework_overview}
\end{figure}

\begin{figure*}[t]
    \centering
    \includegraphics[width=\textwidth]{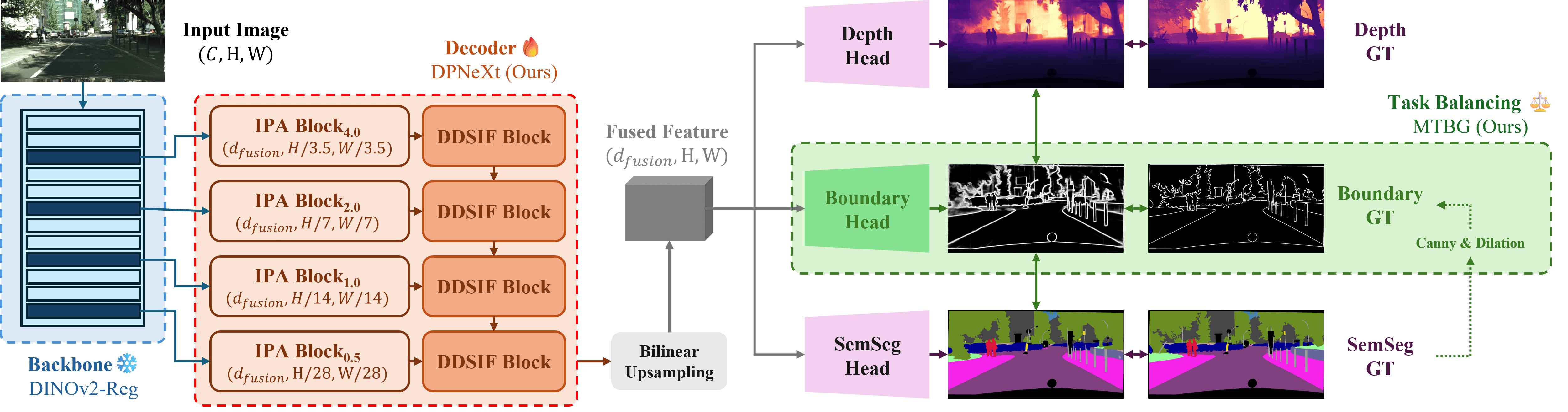}
    \caption[Detailed Architecture of the Proposed MTL Framework.]{\textbf{Detailed Architecture of the Proposed MTL Framework.} 
    (a) The frozen DINOv2-Reg backbone (\textcolor{blue}{Blue}) extracts robust isotropic feature tokens from the input image. 
    (b) The proposed DPNeXt decoder (\textcolor{orange}{Orange}) aggregates these features using the \textit{Isotropic Projection Adapter} (IPA) for dimension alignment and resolution resizing. Subsequently, the \textit{Dual Depthwise Separable Inverted Fusion} (DDSIF) block efficiently fuses the resulting pyramidal features. 
    (c) Task-specific heads (\textcolor{violet}{Purple}) generate predictions for semantic segmentation and depth estimation. 
    (d) The \textit{Multi-Task Boundary Guidance} (MTBG) strategy (\textcolor{green}{Green}) uses provided boundary labels or derives them from segmentation masks when unavailable, requiring no additional annotations. An auxiliary boundary head encourages geometric consistency during training and is discarded during inference to ensure zero latency overhead.}
    \label{fig:architecture}
\end{figure*}

However, operating separate single-task models multiplies parameters and requires redundant fusion modules. Multi-Task Learning (MTL)~\cite{vandenhende2021multi} addresses this by leveraging shared inductive biases to execute multiple tasks~\cite{caruana1997multitask} and extract unified features within a single architecture~\cite{misra2016cross}.

Despite these advantages, existing MTL frameworks face significant decoding limitations. Recent MTL models adopting modern architectures or VFMs still struggle to balance accuracy and efficiency. Naive decoders yield sub-optimal performance, while transformer-based decoders and multi-scale fusion approaches impose heavy computational burdens. Moreover, custom layers often lack native hardware optimization, resulting in slow real-world inference despite low theoretical FLOPs. Existing task balancing strategies compound this by adding auxiliary branches that increase inference latency, and by outputting disparate features that require further redundant fusion downstream.

To address these limitations, we propose an efficient multi-task framework integrating an off-the-shelf VFM encoder, a lightweight multi-scale feature fusion decoder, and a zero-inference-cost task balancing strategy, as illustrated in Fig.~\ref{fig:framework_overview}. Our framework introduces the DPNeXt decoder for efficient feature fusion and the Multi-Task Boundary Guidance (MTBG) strategy to encourage geometric consistency.

Our main contributions are summarized as follows:
\begin{itemize}
    \item We propose the DPNeXt decoder as a highly efficient alternative to standard Dense Prediction Transformer (DPT)~\cite{ranftl2021vision} architectures. By integrating dual depthwise separable inverted bottlenecks, DPNeXt-S reduces trainable parameters by 78.6\% and completely modularizes the task heads.
    \item We introduce MTBG as a symmetric boundary guidance strategy to mitigate negative inductive transfer. It uses label-derived boundary masks to encourage structural consistency, achieving performance gains without additional annotation or inference overhead.
    \item Extensive evaluations on the Cityscapes and NYUv2 benchmarks demonstrate that our framework achieves the best semantic segmentation and depth estimation performance among the compared methods.
    \item Comprehensive computational efficiency analyses demonstrate the framework's fast inference speed and minimal parameter footprint, supporting its applicability to resource-constrained perception settings.
\end{itemize}

%%%%%%%%%%%%%%%%%%%%%%%%%%%%%%%%%%%%%%%%%%%%%%%%%%%%%%%%%%%%%%%%%

\section{RELATED WORKS}

\subsection{Multi-Task Learning for Dense Prediction}
MTL frameworks unify diverse dense prediction tasks. Recent approaches such as InvPT++~\cite{ye2024invpt++} and TaskPrompter~\cite{ye2023taskprompter} rely on global attention mechanisms, incurring computational overheads that can limit real-time deployment. To address this, alternative architectures have emerged, including linear State Space Models (SSMs) in MTMamba~\cite{lin2024mtmamba}, a mixture of low-rank experts in MLoRE~\cite{yang2024multi}, adversarial fine-tuning in SwinMTL~\cite{taghavi2024swinmtl}, and window-based attention in M2H~\cite{udugama2025m2h}. Despite their theoretical efficiency, these methods introduce practical bottlenecks. Custom scan operations and dynamic routing can introduce practical runtime bottlenecks, while adversarial training induces instability, and cross-task modules create structural coupling. In contrast, our approach uses pure-CNN operations that are natively optimized on common hardware platforms, supporting fast inference in resource-constrained environments without unstable training or cross-task dependencies.

\subsection{Evolution of Image Backbones}
Image backbones have evolved from standard convolutional networks~\cite{he2016deep} to efficiency-focused architectures~\cite{howard2017mobilenets,sandler2018mobilenetv2,tan2019efficientnet}, introducing residual learning~\cite{he2016deep}, depthwise separable convolutions~\cite{howard2017mobilenets}, and inverted residual blocks~\cite{liu2022convnet, sandler2018mobilenetv2}. Multi-scale feature refinement~\cite{lin2017refinenet} has further aided in preserving essential spatial details. Concurrently, Vision Transformers (ViTs)~\cite{dosovitskiy2020image} have demonstrated superior scalability and long-range dependency modeling, with hierarchical variants~\cite{liu2021swin} reintroducing inductive biases to handle varying resolutions. Parallel to these architectural shifts, recent perception models increasingly rely on VFMs, notably the DINO series~\cite{caron2021emerging,oquab2023dinov2}. Specifically, DINOv2 provides robust semantic features, and its register-augmented variant~\cite{darcet2024vision} mitigates high-norm artifacts.

Drawing on these dual advancements, our framework adopts the register-augmented DINOv2 as the foundational image backbone. While the backbone handles robust feature extraction, our proposed DPNeXt leverages depthwise separable convolutions and inverted bottleneck structures to redesign RefineNet-style feature fusion blocks. Inspired by ConvNeXt~\cite{liu2022convnet}, which modernized pure convolutional architectures to outperform Vision Transformers, DPNeXt adapts efficient convolutional design principles to ViT-based multi-task dense prediction through a streamlined pure-CNN feature fusion architecture.

\subsection{Feature Fusion Strategies}
Feature fusion integrates spatial details with semantic context. Traditional convolutional networks~\cite{ronneberger2015u,lin2017feature,baheti2020eff} and hierarchical transformers~\cite{liu2022swin, liu2021swin} rely on pyramidal hierarchies, employing artificial channel expansion by assigning smaller dimensions to shallow layers for spatial details and larger dimensions to deep layers for semantic context. However, the dominance of VFMs has shifted the paradigm toward isotropic feature representations, where models like DPT~\cite{ranftl2021vision} explicitly reassemble and fuse constant-resolution tokens. While recent models~\cite{yang2024depth,yang2024depthv2,karypidis2026dino} extensively leverage DPT architectures, many reintroduce CNN-style artificial channel expansion during reassembly. DPNeXt avoids this redundant expansion, maintaining the original DPT strategy of utilizing an isotropic fusion dimension.

Furthermore, explicitly modeling boundary features is known to refine geometric precision. Methods like PIDNet~\cite{xu2023pidnet} and Boundary-Aware MTL~\cite{wang2020boundary} rely on inseparable architectural branches coupled with boundary-aware auxiliary losses, creating cross-task dependencies and inflating inference costs. We address these limitations by employing label-derived boundary guidance with a training-only auxiliary branch, ensuring zero inference overhead while maintaining independent task heads.

%%%%%%%%%%%%%%%%%%%%%%%%%%%%%%%%%%%%%%%%%%%%%%%%%%%%%%%%%%%%%%%%%

\section{METHOD}

\subsection{Framework Formulation}
We formalize our proposed MTL framework, whose detailed architecture is illustrated in Fig.~\ref{fig:architecture}. Let $X^{(i)} \in \mathbb{R}^{3 \times H \times W}$ be the input RGB image tensor for the $i$-th sample in a batch. The pipeline consists of three sequential phases, namely feature extraction, multi-scale reassembly, and task-specific prediction. The overall multi-scale fusion topology strictly follows the established DPT~\cite{ranftl2021vision} architecture.

\subsubsection{Feature Extraction} The frozen backbone extracts a set of isotropic feature tokens. Specifically, we select intermediate features from four layers of the DINOv2-Reg backbone. Let $k \in \{1, 2, 3, 4\}$ denote the feature pyramid level corresponding to backbone layers $l_k \in \{3, 6, 9, 12\}$. The extracted features are defined as follows.
\begin{equation}
    Z_k^{(i)} = \text{DINOv2Reg}_{l_k}(X^{(i)}) \in \mathbb{R}^{d_{\mathrm{backbone}} \times H_p \times W_p},
\end{equation}
where $d_{\mathrm{backbone}}$ is the backbone channel dimension, and 
$H_p = H/14, W_p = W/14$ denote the inherent patch resolution of the ViT.

\subsubsection{Multi-Scale Reassembly} The decoder first projects these diverse features into a unified dimension $d_{\mathrm{fusion}}=256$ and resizes their spatial resolutions using the Isotropic Projection Adapter (IPA). The adapted features are formulated below.
\begin{equation}
    F_k^{(i)} = \text{IPA}_{s_k}(Z_k^{(i)}) \in \mathbb{R}^{d_{\mathrm{fusion}} \times s_k H_p \times s_k W_p},
\end{equation}
where the subscript $s_k \in \{4, 2, 1, 1/2\}$ denotes the spatial resizing scale applied to the $k$-th feature map. Subsequently, the Dual Depthwise Separable Inverted Fusion (DDSIF) blocks recursively aggregate these pyramidal features from deep to shallow layers.
\begin{align}
    P_4^{(i)} &= \text{DDSIF}(F_4^{(i)}), \\
    P_k^{(i)} &= \text{DDSIF}(P_{k+1}^{(i)}, F_k^{(i)}), \quad \text{for } k \in \{3, 2, 1\}.
\end{align}
Each DDSIF block inherently applies a $2\times$ bilinear upsampling prior to its output projection, such that $P_k^{(i)} \in \mathbb{R}^{d_{\mathrm{fusion}} \times 2s_k H_p \times 2s_k W_p}$. The final aggregated feature representation utilized by all task heads is $P_1^{(i)}$.

\subsubsection{Task-Specific Prediction} The shared feature $P_1^{(i)}$ is routed to independent, modular task heads $t \in \{\mathrm{seg}, \mathrm{depth}, \mathrm{bound}\}$. The prediction before the final spatial interpolation is mathematically expressed as follows.
\begin{equation}
    \hat{y}_t^{(i)} = \text{Head}_t(P_1^{(i)}) \in \mathbb{R}^{C_t \times H_f \times W_f},
\end{equation}
where $C_t$ represents the number of output channels for task $t$, and $H_f, W_f$ correspond to the spatial dimensions of $P_1^{(i)}$ (i.e., $H_f = 8H_p$ and $W_f = 8W_p$). Finally, the prediction $\hat{y}_t^{(i)}$ is bilinearly upsampled back to the original input resolution $H \times W$.

\begin{figure}[t]
    \centering
    \includegraphics[width=\linewidth]{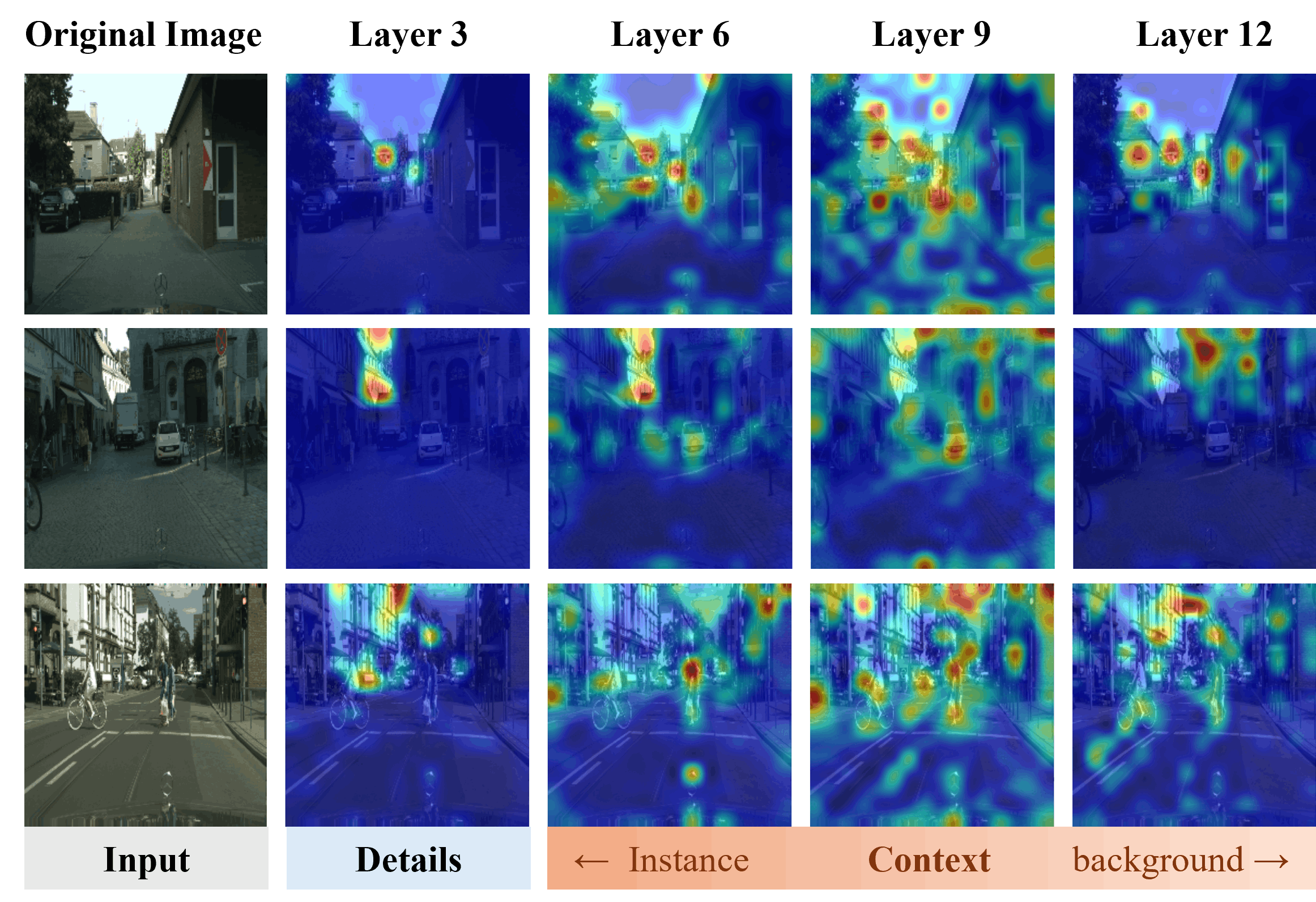}
    \caption[Visualization of DINOv2 Attention Maps.]{\textbf{Visualization of DINOv2 Attention Maps.} 
    We visualize the attention response of the [CLS] token across Layers 3, 6, 9, and 12. Register tokens were removed for spatial readability. 
    (a) \textit{Layer 3} shows sparse and localized attention patterns.
    (b) \textit{Layer 6} begins to highlight object-level regions. 
    (c) \textit{Layer 9} covers broader contextual areas. 
    (d) \textit{Layer 12} exhibits more global scene-level responses.}
    \label{fig:attention_analysis}
\end{figure}

\begin{figure}[t]
    \centering
    \includegraphics[width=\linewidth]{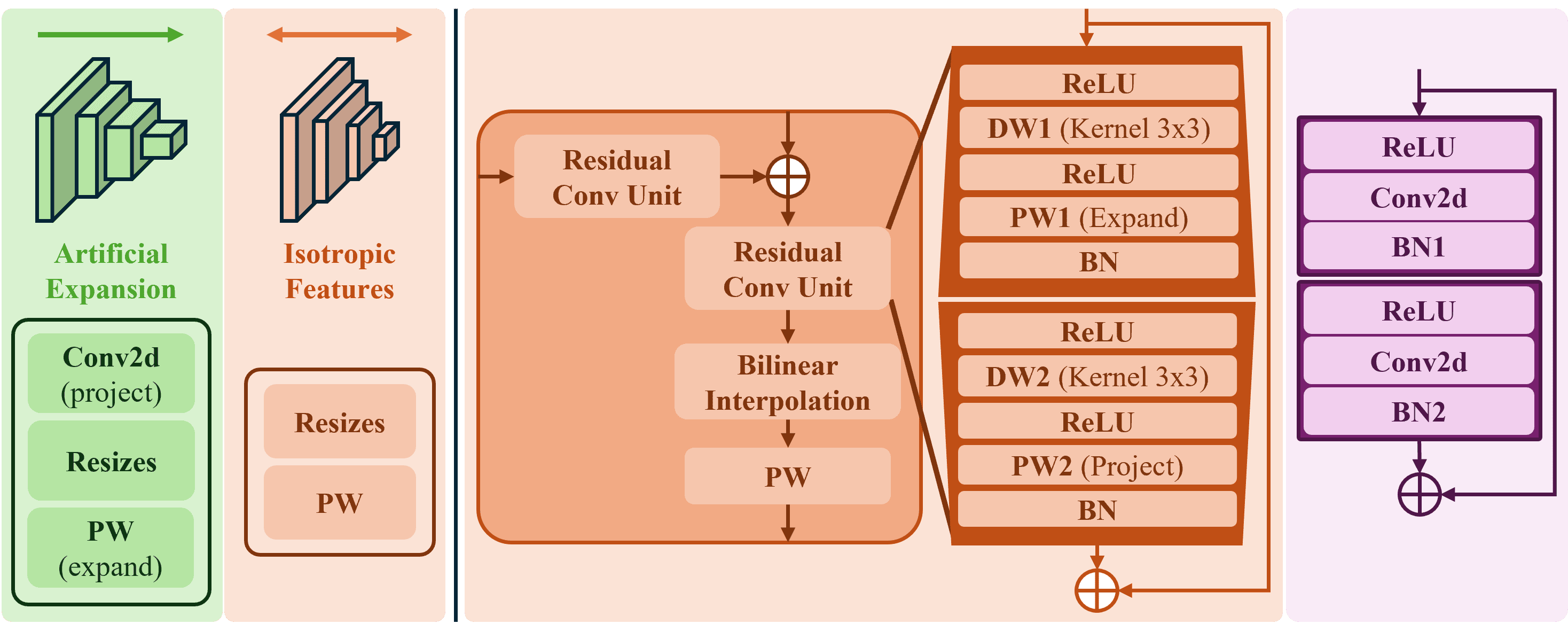}
    \caption{\textbf{Structure of DPNeXt Decoder Components.} 
    \textbf{Left:} We contrast the widely adopted artificial channel expansion strategy (\textcolor{green}{Green}) with the original DPT approach (\textcolor{orange}{Orange}, \textit{Left}) used in our IPA block. By avoiding redundant channel widening with pointwise convolution (PW), IPA projects features into a unified dimension. 
    \textbf{Right:} We redesign the standard DPT residual fusion unit (\textcolor{purple}{Purple}) into our Dual Depthwise Separable Inverted Fusion (DDSIF) block (\textcolor{orange}{Orange}, \textit{Right}). It adapts architectural insights from modern image backbones to support efficient feature fusion with minimal parameter overhead.}
    \label{fig:decoder_components}
\end{figure}

\subsection{DPNeXt}
We introduce DPNeXt, a streamlined decoder composed of the IPA and DDSIF blocks. Using only standard convolutional operations, DPNeXt connects the ViT-based encoder to task heads with low parameter overhead and hardware-friendly inference.

\subsubsection{IPA Block}
Standard DPT implementations often widen feature channels to mimic CNN-style hierarchies. However, the DINOv2 attention analysis in Fig.~\ref{fig:attention_analysis} suggests that semantic grouping emerges early and global context is maintained through self-attention. Therefore, as shown in Fig.~\ref{fig:decoder_components} (left), IPA avoids redundant channel expansion and projects all selected layers into a shared fusion dimension while preserving the resolution hierarchy required for dense prediction.

\subsubsection{DDSIF Block}
As shown in Fig.~\ref{fig:decoder_components} (right), DDSIF reduces the redundancy of standard DPT residual fusion units by replacing dense convolutions with two depthwise separable convolutional stages. The first stage applies depthwise spatial filtering at the fusion dimension and expands the channels by a pointwise convolution with an expansion ratio $r = 2$. The second stage performs depthwise filtering in the expanded space and projects the features back to the fusion dimension using another pointwise convolution. This dual depthwise separable inverted design enables efficient multi-scale aggregation with limited parameter overhead.

\subsection{MTBG}
To mitigate negative transfer, MTBG encourages geometric consistency during training without adding inference latency. Boundary masks are obtained from provided boundary annotations when available. Otherwise, they are derived from semantic segmentation labels using Canny edge detection followed by dilation, following the protocol of PIDNet~\cite{xu2023pidnet}.

\subsubsection{Auxiliary Boundary Loss}
We dynamically compute class weights for each batch to address spatial imbalance. Let $N_{+}$ and $N_{-}$ be the number of positive and negative pixels, respectively, and $N_{\mathrm{total}} = N_{+} + N_{-}$. The weighting factors are $w_{+} = N_{-} / N_{\mathrm{total}}$ and $w_{-} = N_{+} / N_{\mathrm{total}}$. The boundary loss is defined as follows.
\begin{multline}
    \mathcal{L}_{\mathrm{bound}}
    = - \frac{1}{N_{\mathrm{total}}} \sum_{p}
    \big[
    w_{+} y_p \log\big(\sigma(\hat{b}_p)\big) \\
    + w_{-} (1-y_p)
    \log\big(1-\sigma(\hat{b}_p)\big)
    \big].
\end{multline}
where $\hat{b}_p$ is the predicted boundary logit, $y_p$ is the boundary target at pixel $p$, and $\sigma(\cdot)$ is the sigmoid function. 

\subsubsection{Boundary-Aware Loss}
We implement a general boundary-aware loss that applies a base loss only to boundary regions. Let $\mathcal{M}_{\mathrm{bound}}$ denote the binary boundary mask, either provided directly or derived from semantic labels when boundary annotations are unavailable. The Boundary-Aware Segmentation (BAS) and Boundary-Aware Depth (BAD) losses are formulated as dual objectives.
\begin{align}
    \mathcal{L}_{\mathrm{bas}} &= 
    \mathcal{L}_{\mathrm{seg}}(\hat{y}_{\mathrm{seg}}, y_{\mathrm{seg}};\mathcal{M}_{\mathrm{bound}}), \\
    \mathcal{L}_{\mathrm{bad}} &= 
    \mathcal{L}_{\mathrm{depth}}(\hat{y}_{\mathrm{depth}}, y_{\mathrm{depth}};\mathcal{M}_{\mathrm{bound}}),
\end{align}
where $\mathcal{M}_{\mathrm{bound}}$ restricts each loss to valid boundary pixels.

\subsection{Loss Functions}
To optimize our multi-task framework, we employ specific objective functions for each task head.

\subsubsection{Semantic Segmentation Loss}
To address class imbalance in dense prediction, we adopt a pixel-wise OHEM variant of cross-entropy inspired by~\cite{shrivastava2016training}. We select hard pixels $\mathcal{K}$ whose predicted probabilities for the ground-truth class fall below a threshold $\tau$.
\begin{align}
    \mathcal{L}_{\mathrm{seg}} 
    &= - \frac{1}{|\mathcal{K}|} 
    \sum_{p \in \mathcal{K}} \log(\hat{p}_{p, y_p}), \\
    \mathcal{K} 
    &= \{p \mid \hat{p}_{p, y_p} < \tau \}.
\end{align}

\subsubsection{Depth Estimation Loss}
For absolute depth recovery, we utilize the Scale-Invariant Logarithmic (SiLog) loss~\cite{eigen2014depth}.
\begin{equation}
    \mathcal{L}_{\mathrm{depth}} = \sqrt{\frac{1}{N} \sum_{p} \Delta_p^2 - \frac{\lambda_{\mathrm{silog}}}{N^2} \left( \sum_{p} \Delta_p \right)^2},
\end{equation}
where $\Delta_p = \log d_p - \log \hat{d}_p$ is the logarithmic difference 
between the ground truth and the prediction. We set $\lambda_{\mathrm{silog}}=0.5$ to balance variance minimization with absolute error penalization. To mitigate severe distribution imbalance, the metric depth $d$ is clipped to $\tilde{d} = \operatorname{clip}(d, d_{\mathrm{min}}, d_{\mathrm{max}})$ and log-normalized to $d_{\mathrm{norm}} \in [0, 1]$ following SwinMTL~\cite{taghavi2024swinmtl}. While SwinMTL applied this normalization exclusively within its auxiliary critic network, we integrate it directly into our primary MTL generator network. The normalized depth is computed below.
\begin{equation}
    d_{\mathrm{norm}} = \frac{\log(\tilde{d} / d_{\mathrm{min}})}{\log(d_{\mathrm{max}} / d_{\mathrm{min}})}.
\end{equation}
The model outputs this normalized depth, which is inversely transformed back to the metric scale before computing the SiLog loss.

\subsubsection{Surface Normal Loss}
For the NYUv2 dataset exclusively, we incorporate surface normal estimation as an additional auxiliary task to further refine geometric representations. This is optimized using a simple $\mathcal{L}_1$ loss between the predicted and ground truth normal vectors. Similar to the boundary head, this branch is completely discarded during inference.

\subsubsection{Total Loss}
The final objective function is a linear combination of the primary task losses and the auxiliary MTBG losses, fine-tuned from initial ratios derived via homoscedastic uncertainty weighting~\cite{kendall2018multi}.
\begin{equation}
\begin{split}
    \mathcal{L}_{\mathrm{total}} &= \lambda_{\mathrm{seg}} \mathcal{L}_{\mathrm{seg}} + \lambda_{\mathrm{depth}} \mathcal{L}_{\mathrm{depth}} + \lambda_{\mathrm{bound}} \mathcal{L}_{\mathrm{bound}} \\
    &\quad + \lambda_{\mathrm{bas}} \mathcal{L}_{\mathrm{bas}} + \lambda_{\mathrm{bad}} \mathcal{L}_{\mathrm{bad}}.
\end{split}
\end{equation}
For NYUv2 experiments, the auxiliary surface normal loss is simply appended to this total objective. The loss weights are reported together with the training settings in Sec.~\ref{sec:experimental_setup}.

\begin{table}[t]
    \centering
    \caption[Performance Comparison with SOTA MTL Methods on Cityscapes Validation Set]{Performance Comparison with SOTA MTL Methods \\ on Cityscapes Validation Set}
    \label{tab:cityscape_benchmark}
    \setlength{\tabcolsep}{3.5pt} 
    \resizebox{\columnwidth}{!}{%
    \begin{tabular}{l c c c c}
        \toprule
        \multirow{2}{*}{\textbf{Model}} & \multirow{2}{*}{\textbf{\shortstack{Params\\(Train.) [M]} $\downarrow$}} & \textbf{SemSeg} & \textbf{Depth} & \textbf{Total} \\
        \cmidrule(lr){3-3} \cmidrule(lr){4-4} \cmidrule(lr){5-5}
         & & \textbf{mIoU} $\uparrow$ [\%] & \textbf{RMSE} $\downarrow$ [m] & \textbf{JPS} $\uparrow$ \\
        \midrule
        PAD-Net~\cite{xu2018pad, lopes2023cross} & -$^\ddagger$ & 70.23 & 6.78 & 0.809 \\
        MTL~\cite{vandenhende2021multi, lopes2023cross} & -$^\ddagger$ & 70.43 & 6.80 & 0.810 \\
        DenseMTL~\cite{lopes2023cross} & 87.4 (87.4) & 74.95 & 6.65 & 0.833 \\
        3-ways~\cite{hoyer2021three, lopes2023cross} & 87.2 (87.2) & 75.00 & 6.53 & 0.834 \\
        SwinMTL~\cite{taghavi2024swinmtl} & 87.4 (87.4) & 76.41 & 6.32 & 0.843 \\
        DPT-DINOv2-S~\cite{oquab2023dinov2} & 52.2 (30.2) & 75.40 & 5.52 & 0.843 \\
        M2H~\cite{udugama2025m2h} & -$^\dagger$ & 77.60$^\dagger$ & 6.10$^\dagger$ & 0.850$^\dagger$ \\
        \midrule
        DPNeXt-S (Ours) & 28.5 (6.5) & \underline{78.32} & \underline{5.45} & \underline{0.858} \\
        DPNeXt-B (Ours) & 93.4 (6.8) & \textbf{79.64} & \textbf{4.95} & \textbf{0.867} \\
        \bottomrule
        \multicolumn{5}{p{0.98\columnwidth}}{\rule{0pt}{2.5ex}\scriptsize \textbf{Bold} and \underline{underline} indicate the best and second-best performances, respectively.} \\
        \multicolumn{5}{p{0.98\columnwidth}}{\scriptsize $^\dagger$Official code and checkpoints from the original authors are unavailable.} \\
        \multicolumn{5}{p{0.98\columnwidth}}{\scriptsize $^\ddagger$Code corresponding to reported values is unavailable.}
    \end{tabular}%
    }
\end{table}

\begin{table}[t]
    \centering
    \caption[Performance Comparison with SOTA MTL Methods on NYUv2 Test Set]{Performance Comparison with SOTA MTL Methods \\ on NYUv2 Test Set}
    \label{tab:nyuv2_benchmark}
    \setlength{\tabcolsep}{3.5pt} 
    \resizebox{\columnwidth}{!}{%
    \begin{tabular}{l c c c c}
        \toprule
        \multirow{2}{*}{\textbf{Model}} & \multirow{2}{*}{\textbf{\shortstack{Params\\(Train.) [M]} $\downarrow$}} & \textbf{SemSeg} & \textbf{Depth} & \textbf{Total} \\
        \cmidrule(lr){3-3} \cmidrule(lr){4-4} \cmidrule(lr){5-5}
         & & \textbf{mIoU} $\uparrow$ [\%] & \textbf{RMSE} $\downarrow$ [m] & \textbf{JPS} $\uparrow$ \\
        \midrule
        InvPT~\cite{ye2022inverted} & 402.1 (402.1) & 53.56 & 0.5183 & 0.742 \\
        TaskPrompter~\cite{ye2023taskprompter} & 359.5 (359.5) & 55.30 & 0.5152 & 0.751 \\
        MTMamba~\cite{lin2024mtmamba} & 252.27 (252.27) & 55.82 & 0.5066 & 0.754 \\
        MLoRE~\cite{yang2024multi} & 550.8 (550.8) & 55.96 & 0.5076 & 0.754 \\
        MTMamba++~\cite{lin2025mtmamba++} & 258.6 (258.6) & 57.01 & 0.4818 & 0.761 \\
        SwinMTL~\cite{taghavi2024swinmtl} & 87.4 (87.4) & 58.14$^\dagger$ & 0.5179$^\dagger$ & 0.765$^\dagger$ \\
        M2H-Small~\cite{udugama2025m2h} & 33.7 (11.6) & 58.05$^\ddagger$ & 0.4365$^\ddagger$ & 0.768$^\ddagger$ \\
        M2H~\cite{udugama2025m2h} & 81.54 (59.48) & \underline{61.54}$^\ddagger$ & \underline{0.4196}$^\ddagger$ & \underline{0.787}$^\ddagger$ \\
        \midrule
        DPNeXt-S (Ours) & 28.9 (6.8) & 57.15 & 0.4783 & 0.762 \\
        DPNeXt-B (Ours) & 93.8 (7.2) & \textbf{62.01} & \textbf{0.4168} & \textbf{0.789} \\
        \bottomrule
        \multicolumn{5}{p{0.98\columnwidth}}{\rule{0pt}{2.5ex}\scriptsize \textbf{Bold} and \underline{underline} indicate the best and second-best performances, respectively.} \\
        \multicolumn{5}{p{0.98\columnwidth}}{\scriptsize $^\dagger$ Official code and checkpoints from the original authors are unavailable.} \\
        \multicolumn{5}{p{0.98\columnwidth}}{\scriptsize $^\ddagger$ Official checkpoint results are lower than the reported values.} \\
    \end{tabular}%
    }
\end{table}

\section{EXPERIMENTS}

\subsection{Experimental Setup}
\label{sec:experimental_setup}
We evaluate our framework on two standard dense prediction benchmarks. 
The Cityscapes dataset~\cite{cordts2016cityscapes} provides urban driving scenes, 
with 2,975 training and 500 validation images. We evaluate 19 semantic 
categories and depth estimation restricted to the $[0.001, 80.0]$m range using 
CREStereo~\cite{li2022practical} depth maps, following prior MTL protocols~\cite{taghavi2024swinmtl,udugama2025m2h}. The NYUv2 dataset~\cite{silberman2012indoor} comprises 795 training and 654 testing indoor RGB-D pairs, evaluated on a 40-class semantic configuration and a $[0.001, 10.0]$m depth range.

Main benchmark models are trained for 400 epochs with a total batch size of 8 using the AdamW optimizer with $(\beta_1,\beta_2)=(0.9,0.999)$ and weight decay $10^{-4}$, 
excluding bias and normalization parameters. The base learning rate is $10^{-3}$, 
with linear warm-up for the first 5\% of total updates from a start factor of 
$10^{-3}$, followed by polynomial decay with power 1.0. We use color augmentations, 
including brightness/contrast adjustment, gamma correction, and HSV perturbations, 
together with horizontal flipping and random scaling. The OHEM threshold is set 
to $\tau=0.9$ with a minimum kept-pixel ratio of $1/8$. For Cityscapes, the loss 
weights $(\lambda_{\mathrm{seg}},\lambda_{\mathrm{depth}},\lambda_{\mathrm{bound}})$ 
are set to $(1.0,1.0,50.0)$. For NYUv2, $(\lambda_{\mathrm{seg}},\lambda_{\mathrm{depth}},\lambda_{\mathrm{bound}},
\lambda_{\mathrm{normal}})$ is set to $(1.5,1.0,50.0,10.0)$ for DPNeXt-S 
and $(1.0,2.0,50.0,10.0)$ for DPNeXt-B. The boundary-aware loss weights are 
fixed at $\lambda_{\mathrm{bas}}=\lambda_{\mathrm{bad}}=0.4$.

Main benchmark models were trained on an NVIDIA RTX 3090 GPU. 
Models used in ablation studies were trained on a workstation equipped with four NVIDIA Titan RTX GPUs for faster experimental iteration, using 160 epochs and a total batch size of 32. Accuracy evaluation and inference speed measurement were conducted on an NVIDIA RTX 2080 Laptop GPU as a proxy for a resource-constrained deployment setting.

\begin{table}[t]
    \centering
    \caption[Computational Efficiency and Inference Speed Comparison]{Computational Efficiency and Inference Speed Comparison}
    \label{tab:efficiency_benchmark}
    \setlength{\tabcolsep}{3.5pt} 
    \resizebox{\columnwidth}{!}{%
    \begin{tabular}{l c c c}
        \toprule
        \textbf{Model} & \textbf{Params (Train.) [M]} $\downarrow$ & \textbf{GFLOPs [G]} $\downarrow$ & \textbf{FPS [Hz]} $\uparrow$ \\
        \midrule
        MLoRE~\cite{yang2024multi} & 550.77 (550.77) & 527.53 & 5.63 \\
        M2H~\cite{udugama2025m2h} & 81.54 (59.48) & 843.71 & 6.18 \\
        TaskPrompter~\cite{ye2023taskprompter} & 359.54 (359.54) & 586.55 & 14.63 \\
        MTMamba++~\cite{lin2025mtmamba++} & 258.6 (258.6) & 104.81 & 17.63 \\
        SwinMTL~\cite{taghavi2024swinmtl} & 87.39 (87.39) & \underline{59.37} & 19.74 \\
        M2H-Small~\cite{udugama2025m2h} & \underline{33.7} (11.64) & \textbf{51.6} & 20.72 \\
        \midrule
        DPNeXt-B (Ours) & 93.81 (\underline{7.22}) & 131.83 & \underline{36.49} \\
        \textbf{DPNeXt-S (Ours)} & \textbf{28.89} (\textbf{6.83}) & 98.24 & \textbf{51.02} \\
        \bottomrule
        \multicolumn{4}{p{0.98\columnwidth}}{\rule{0pt}{2.5ex}\scriptsize \textbf{Bold} and \underline{underline} indicate the best and second-best performances, respectively.} \\
        \multicolumn{4}{p{0.98\columnwidth}}{\scriptsize Input image resolution is fixed to $224 \times 224$.}
    \end{tabular}%
    }
\end{table}

\begin{table}[t]
    \centering
    \caption[Ablation Studies on DPNeXt Architecture and MTBG Strategy]{Ablation Studies on DPNeXt Architecture \\ and MTBG Strategy}
    \label{tab:ablation_study}
    \setlength{\tabcolsep}{2.5pt}
    \resizebox{\columnwidth}{!}{%
    \begin{tabular}{@{} c c c c c c c @{}}
        \toprule
        % DPNeXt Architecture
        \multicolumn{3}{l}{\multirow{2}{*}{\textbf{Method}}} & \multirow{2}{*}{\textbf{\shortstack{Params\\(Train.) [M]}} $\downarrow$} & \textbf{SemSeg} & \textbf{Depth} & \textbf{Total} \\
        \cmidrule(lr){5-5} \cmidrule(lr){6-6} \cmidrule(lr){7-7}
        \multicolumn{3}{l}{} & & \textbf{mIoU} $\uparrow$ [\%] & \textbf{RMSE} $\downarrow$ [m] & \textbf{JPS} $\uparrow$ \\
        \midrule
        \multicolumn{3}{l}{Standard DPT} & 52.236 (30.177) & 68.261 & 7.220 & 0.796 \\
        \multicolumn{3}{l}{+ IPA} & 37.412 (15.353) & 68.229 & 7.080 & 0.797 \\
        \multicolumn{3}{l}{+ DDSIF: Dual DSConv} & 27.062 (5.004) & \underline{68.952} & \underline{6.839} & \underline{0.802} \\
        \multicolumn{3}{l}{+ DDSIF: Inverted Bottleneck} & 28.131 (6.073) & \textbf{69.660} & \textbf{6.803} & \textbf{0.806} \\
        \midrule
        \midrule
        % MTBG Strategy
        \multicolumn{3}{c}{\textbf{MTBG Strategy}} & \multirow{2}{*}{\textbf{\shortstack{Params\\(Train.) [M]}}} & \textbf{SemSeg} & \textbf{Depth} & \textbf{Total} \\
        \cmidrule(lr){1-3} \cmidrule(lr){5-5} \cmidrule(lr){6-6} \cmidrule(lr){7-7}
        \makebox[1.15cm][c]{\textbf{OHEM}} & \makebox[1.15cm][c]{\textbf{BAS}} & \makebox[1.15cm][c]{\textbf{BAD}} & & \textbf{mIoU} $\uparrow$ [\%] & \textbf{RMSE} $\downarrow$ [m] & \textbf{JPS} $\uparrow$ \\
        \midrule
        \makebox[1.15cm][c]{\ding{51}} & \makebox[1.15cm][c]{\ding{55}} & \makebox[1.15cm][c]{\ding{55}} & 28.131 (6.073) & 70.658 & 6.998 & 0.810 \\
        \makebox[1.15cm][c]{\ding{51}} & \makebox[1.15cm][c]{\ding{55}} & \makebox[1.15cm][c]{\ding{51}} & 28.514 (6.456) & 70.911 & \textbf{6.805} & \underline{0.812} \\
        \makebox[1.15cm][c]{\ding{51}} & \makebox[1.15cm][c]{\ding{51}} & \makebox[1.15cm][c]{\ding{55}} & 28.514 (6.456) & \underline{71.066} & 7.083 & 0.811 \\
        \makebox[1.15cm][c]{\ding{51}} & \makebox[1.15cm][c]{\ding{51}} & \makebox[1.15cm][c]{\ding{51}} & 28.514 (6.456) & \textbf{71.391} & \underline{6.864} & \textbf{0.814} \\
        \bottomrule
        \multicolumn{7}{@{}p{0.95\columnwidth}@{}}{\rule{0pt}{2.5ex}\scriptsize \textbf{Bold} and \underline{underline} indicate the best and second-best performances, respectively.}
    \end{tabular}%
    }
\end{table}

For Cityscapes, some prior methods are not evaluated at full image resolution, and exact reproduction is not always possible. To ensure a fair and conservative comparison, we report the best available baseline results from the original published results, third-party reproductions, and our own empirical evaluations. We also train an additional strong baseline, DPT-DINOv2-S, by combining a DINOv2-S backbone with the standard DPT decoder based on a public implementation~\cite{karypidis2026dino}. We train our DPNeXt models and DPT-DINOv2-S using $518 \times 518$ random crops. We evaluate them on full-resolution $1024 \times 2048$ images using single-scale sliding-window inference with a $518 \times 518$ crop, an overlap ratio of 0.5, and Gaussian blending only for stitching overlapping predictions. No test-time augmentation, such as horizontal flipping or multi-scale inference, is used.

For NYUv2, we follow the commonly used protocol adopted by InvPT~\cite{ye2022inverted}. Models are trained with random crops and evaluated by resizing images to $448 \times 576$ for a single forward pass, without sliding-window stitching, Gaussian blending, horizontal flipping, or multi-scale inference.

To summarize the joint performance of semantic segmentation and depth estimation, 
we define the Joint Performance Score (JPS). For our primary two-task evaluation, 
we scale the mIoU term to $[0,1]$ and normalize depth RMSE by the dataset depth 
range $(d_{\mathrm{max}} - d_{\mathrm{min}})$:
\begin{equation}
\mathrm{JPS} = \frac{1}{2} \left(
\frac{\mathrm{mIoU}}{100} +
\max \left(0, 1 - \frac{\mathrm{RMSE}}{d_{\mathrm{max}} - d_{\mathrm{min}}} \right)
\right).
\end{equation}
This gives equal weight to semantic segmentation and depth estimation.

\subsection{Quantitative Results}

Tables~\ref{tab:cityscape_benchmark} and \ref{tab:nyuv2_benchmark} summarize the 
quantitative comparisons on Cityscapes and NYUv2. On Cityscapes, DPNeXt-B achieves 
the best overall performance among the compared methods with a JPS of 0.867. 
Notably, DPNeXt-S also outperforms prior baselines while using fewer than 7M 
trainable parameters, achieving a JPS of 0.858. This shows that the proposed 
decoder improves the accuracy-efficiency trade-off of ViT-based dense prediction.

On NYUv2, DPNeXt-S remains competitive despite its small trainable parameter 
count. DPNeXt-B further achieves the best semantic segmentation and depth estimation 
performance among the compared methods, with a JPS of 0.789, while requiring substantially fewer trainable parameters than large-scale MTL baselines such as InvPT, TaskPrompter, MTMamba++, and MLoRE.

Overall, these results suggest that DPNeXt scales effectively across backbone 
sizes and dense prediction benchmarks. By leveraging off-the-shelf VFMs without 
modifying the backbone architecture, DPNeXt provides an efficient alternative to 
the standard DPT decoder with a lightweight trainable parameter footprint.

\subsection{Qualitative Results}

\begin{figure*}[t]
    \centering
    \includegraphics[width=0.95\textwidth]{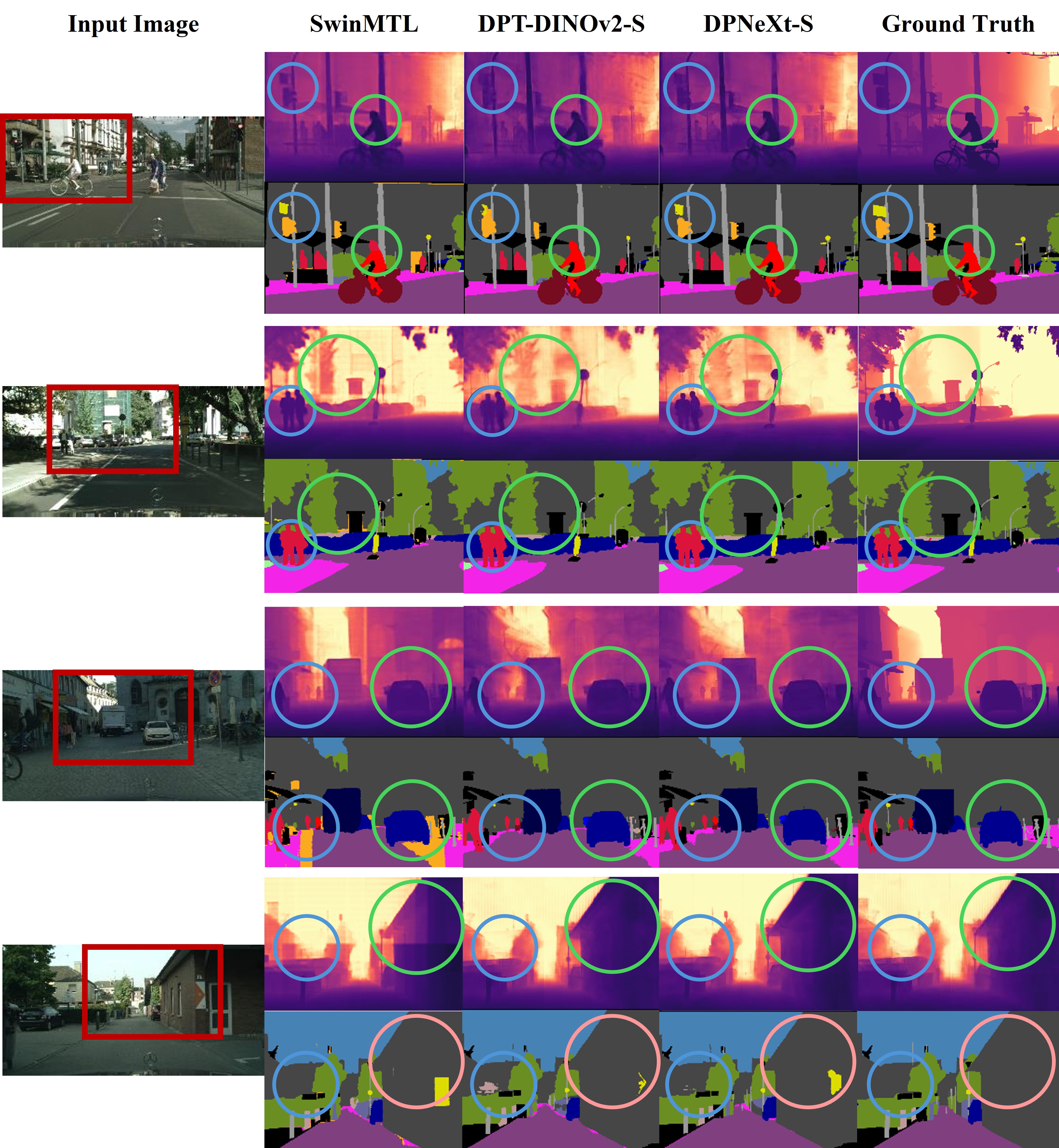}
    \caption{\textbf{Qualitative Results on the Cityscapes Dataset.} We compare the predictions of SwinMTL, DPT-DINOv2-S, and our DPNeXt-S against the GT across diverse urban scenes. For enhanced visualization, specific regions are cropped and magnified. The original crop regions are indicated by red rectangles on the RGB input images. Blue and green circles highlight areas where DPNeXt-S demonstrates superior structural accuracy compared to the baselines. Conversely, red circles indicate failure cases of our model.}
    \label{fig:qualitative_results}
\end{figure*}

Figure~\ref{fig:qualitative_results} presents a qualitative comparison of DPNeXt-S 
against our reproduced SwinMTL and DPT-DINOv2-S baselines across diverse urban 
scenes from the Cityscapes dataset. Across most scenarios, DPNeXt-S produces sharper 
depth maps and more accurate semantic segmentation masks. Notably, SwinMTL relies on a window-based Swin Transformer backbone, which can introduce visible window-shaped artifacts, as observed in the depth predictions of the fourth scenario. In contrast, such artifacts are not observed in DPNeXt-S, which uses standard convolutional fusion blocks without window-partitioned attention in the decoder.

Furthermore, DPNeXt-S shows improved boundary delineation compared to DPT-DINOv2-S. 
For instance, in the first scene, it better preserves the outlines of the cyclist 
and traffic light. These visual examples support the effectiveness of the proposed 
MTBG strategy in encouraging structural consistency and refining geometric boundaries.

\subsection{Computational Efficiency}

To assess the feasibility of efficient inference in resource-constrained settings, 
we evaluate computational efficiency against baseline models in 
Table~\ref{tab:efficiency_benchmark}. We report parameter counts, GFLOPs, and 
actual inference speed measured on the RTX 2080 Laptop GPU. Although DPNeXt-S 
has higher GFLOPs than some lightweight baselines, it achieves the fastest 
measured inference speed. This indicates that theoretical GFLOPs do not fully 
capture practical hardware efficiency. By using standard convolutional operations 
without complex custom layers, DPNeXt provides hardware-friendly inference with 
a small trainable parameter footprint.

\subsection{Ablation Studies}

\subsubsection{From DPT to DPNeXt}
Table~\ref{tab:ablation_study} outlines the architectural evolution from the 
standard DPT to our efficient DPNeXt design. The baseline DPT decoder contains 
30.2M trainable parameters. Introducing the IPA module removes artificial channel 
expansion and substantially reduces the trainable parameter count while maintaining 
performance. Configuring the DDSIF block with a dual depthwise separable structure 
further reduces parameters and improves overall performance. Finally, adding the 
inverted bottleneck improves feature aggregation with only a small increase in 
parameters. We set the expansion ratio to two to avoid parameter bloat while 
retaining efficient multi-scale fusion.

\subsubsection{Effects of MTBG}
The second half of Table~\ref{tab:ablation_study} evaluates our MTBG strategy. 
We analyze the impact of asymmetric boundary guidance by applying BAS and BAD 
separately. The BAD-only variant improves depth estimation from 6.998 to 6.805 
RMSE and also slightly improves semantic segmentation. In contrast, the BAS-only 
variant improves semantic segmentation from 70.658 to 71.066 mIoU but degrades 
depth estimation from 6.998 to 7.083 RMSE, indicating a task trade-off. The full 
MTBG formulation applies boundary guidance to both tasks and achieves the best 
aggregate JPS, requiring only a marginal increase of 0.38M trainable parameters 
with zero inference cost.

\section{CONCLUSION}
This study presented an efficient multi-task dense prediction framework integrating a frozen VFM backbone, the DPNeXt decoder, and the MTBG strategy. DPNeXt reduces trainable parameters through fusion-centric decoding, while MTBG mitigates negative inductive transfer with zero inference cost. Experiments on Cityscapes and NYUv2 demonstrate a strong accuracy-efficiency trade-off, including the best semantic segmentation and depth estimation performance among the compared methods and fast inference with a lightweight trainable parameter footprint. These results suggest that DPNeXt can serve as a practical dense-prediction framework for resource-constrained robotics. Future work will extend this single-frame dense prediction framework toward mobile deployment for streaming and open-vocabulary perception with feature-space analysis, as well as downstream embodied validation for dynamic tasks such as end-to-end autonomous driving and racing.

\bibliographystyle{IEEEtran}
\bibliography{refs}

\addtolength{\textheight}{-12cm}   
% This command serves to balance the column lengths
% on the last page of the document manually. It shortens
% the textheight of the last page by a suitable amount.
% This command does not take effect until the next page
% so it should come on the page before the last. Make
% sure that you do not shorten the textheight too much.

%%%%%%%%%%%%%%%%%%%%%%%%%%%%%%%%%%%%%%%%%%%%%%%%%%%%%%%%%%%%%%%%%%%%%%%%%%%%%%%%
% \section*{APPENDIX}
% \section*{ACKNOWLEDGMENT}
\end{document}